\documentclass[12pt]{article}
\usepackage{amsmath}
\usepackage{graphicx}
\usepackage{hyperref}
\usepackage{algorithm}
\usepackage{algpseudocode}
\usepackage{fancyhdr}
\usepackage{array} 
\usepackage{booktabs} 

\title{A Method for Building Large Language Models with Predefined KV Cache Capacity}
\author{
	Zhonghua Yi\textsuperscript{*}, Ge Niu, Lei Wang, Wei Tang, Liqiu Zhang \\
	DEEPAI Co., Ltd., Beijing \\
	\texttt{zhyi@dpai.com\textsuperscript{*}}
}
\date{October 16, 2024}

\begin{document}
	
	\maketitle
	
	\begin{abstract}
		This paper introduces a novel approach, the Bounded-Cache Transformer (BCT), for building large language models with a predefined Key-Value (KV) cache capacity. The BCT addresses the excessive memory consumption issue in traditional KV caches by implementing a bounded-length KV cache, which is particularly suitable for the attention layers in Transformer decode-only architectures. By dynamically updating the key-value vector sequences, the BCT achieves efficient inference within limited cache capacity, significantly reducing memory usage while maintaining model performance and system throughput. Experimental results demonstrate that the BCT significantly reduces memory usage while maintaining the model's inference quality, offering a new solution for efficient inference in large language models.
	\end{abstract}
	
	\section{Introduction}
	In recent years, large-scale pre-trained models such as BERT \cite{devlin2018bert} and GPT \cite{radford2019language} have achieved significant progress in natural language processing (NLP) tasks, especially in text generation, machine translation, and sentiment analysis. However, these models typically contain hundreds of millions or even billions of parameters, leading to substantial memory and computational resource demands during inference. Particularly, traditional KV cache mechanisms expand continuously with increasing context length, resulting in excessive memory consumption and negatively impacting model performance and system stability \cite{vaswani2017attention}.
	
	To address this issue, this paper proposes a method for building large language models with predefined KV cache capacity. This method sets a fixed cache capacity and dynamically updates the key-value vector sequences, achieving efficient and stable inference. Specifically, the proposed method improves traditional KV cache mechanisms in the following ways:
	1. Fixed Cache Capacity: By setting a fixed cache length, it limits memory usage and avoids uncontrolled growth with increasing context length.
	2. Dynamic Update Mechanism: By continuously updating the key-value vector sequences, it ensures that the information in the cache remains up-to-date, thus maintaining the model's inference quality.
	3. Efficient Inference: By optimizing cache management and update mechanisms, it enhances inference speed and system throughput.
	
	\section{Related Work}
	\subsection{Streaming LLM}
	Streaming LLM is an efficient framework for handling long texts and stream data, focusing on optimizing Key-Value (KV) processing to adapt to long text and stream applications. Streaming LLM retains attention pools, combines recent token rolling caches, and discards unnecessary intermediate tokens, achieving efficient KV cache management. Its advantages include efficient handling of long texts, no need for fine-tuning, improved inference speed, and maintaining accuracy. However, it also has limitations such as context dependency and pre-training window restrictions \cite{xiao2023efficient}.
	
	\subsection{Low-Rank Embedding with Sparse Strategy (LESS)}
	LESS is a method combining sparse strategies and low-rank embeddings to optimize Key-Value (KV) caches, improving the efficiency and performance of large language models (LLMs) when handling long texts and stream data \cite{dong2024get}. LESS synthesizes sparse KV policies and low-rank states, bridging performance gaps, especially in tasks where sparse algorithms show weaknesses. It approximates residuals in sparse caches using low-rank methods, thereby enhancing cache efficiency without significantly increasing memory usage. LESS offers advantages such as performance improvement, low memory occupancy, and ease of integration. It has shown impressive performance across multiple datasets, models, and sparse policies, achieving near-full cache model performance in tasks like language modeling and classification.
	
	\subsection{Neural Turing Machine (NTM)}
	NTM is an enhanced neural network with external storage matrices capable of reading from and writing to external storage. They are designed to simulate the behavior of Turing machines and handle complex tasks requiring access and manipulation of memory \cite{graves2014neural}. This paper draws inspiration from NTM, introducing fixed-length KV caches and dynamic update mechanisms to ensure that the information in the cache remains up-to-date. This approach maintains the model's inference quality while significantly reducing memory usage and improving system throughput.
	
	\subsection{Comparison of RNN, Transformer, and BCT}
	\label{sec:comparison}
	We compare the characteristics of RNNs, Transformers, and our proposed BCT in Table \ref{tabular:comparison}.
	
	\begin{table}[ht]
		\centering			
		\renewcommand{\arraystretch}{0.8} 
		\begin{tabular}{@{}c>{\scriptsize}c>{\scriptsize}c>{\scriptsize}c>{\scriptsize}c@{}}
			\toprule
			Model & State & Mem & Comp \\
			& Expressiveness & Consumption & Complexity \\
			\midrule
			RNN & Fixed-size vector & Bounded & Linear \\
			Transformer & Unbounded & Unbounded & Quadratic \\
			BCT & Bounded-length vectors & Bounded & Linear \\
			\bottomrule
		\end{tabular}
		\caption{Comparison of RNNs, Transformers, and BCT in terms of state management and computational properties.}
		\label{tabular:comparison}
	\end{table}
	
	\section{Method}
\subsection{Model Structure}
The proposed method is applied to the attention layers in large language models based on the Transformer architecture. The core components of the attention layer include:
- \textbf{Key-Value Vector Sequence} $\mathbf{MV}$: Composed of $ M $ key-value vectors, where $ M $ is a predefined value.
- \textbf{Key Vector Sequence} $\mathbf{MK}$: Composed of $ M $ key vectors, where $ M $ is a predefined value.

Initially, the key-value vector sequence $\mathbf{MV}$ can be initialized to zero or random values, depending on the application requirements.

\subsection{Key-Value Vector Update Mechanism}
\label{subsec:kv_update_mechanism}

The proposed Bounded-Cache Transformer (BCT) employs a mechanism for updating key-value vectors that is central to its operation within the attention layers of the Transformer architecture. This mechanism is designed to optimize the use of a predefined, bounded cache capacity, enhancing both memory efficiency and computational throughput. The update process is detailed as follows:

\begin{enumerate}
	\item \textbf{Mapping of Input Vectors:} The input sequence is initially projected into a higher-dimensional space using learnable matrices $\mathbf{W}_{wq}$ and $\mathbf{W}_{wv}$, generating the write query vector $\mathbf{wq}$ and the write key-value vector $\mathbf{wv}$, respectively. This step is crucial for transforming the input data into a form that can be effectively processed by the attention mechanism.
	
	\item \textbf{Computation of Attention Weights:} The write query vector $\mathbf{wq}$ is then used to compute attention weights across the key vectors $\mathbf{MK}$ stored in the cache. This is achieved through a dot-product operation, followed by a softmax normalization, yielding the attention weights $\mathbf{ww}$. These weights determine the influence of each key vector on the subsequent update.
	
	\item \textbf{Retrieval and Update of Key-Value Cache:} For each input vector, the corresponding key-value vector is retrieved from the cache. If it is the first input ($N = 1$), the initial key-value vector sequence $\mathbf{MV}$ is used; otherwise, the sequence from the previous step ($N-1$) is retrieved. This retrieval process is pivotal for the stateful progression of the model.
	
	\item \textbf{Application of Update Rule:} The update rule is applied by integrating the attention weights $\mathbf{ww}$ with the write key-value vector $\mathbf{wv}$. This integration is mathematically represented as $\mathbf{MV} = \mathbf{MV}' + \mathbf{ww}^T \cdot \mathbf{wv}$, where $\mathbf{MV}'$ is the retrieved key-value vector sequence. This operation effectively updates the cache content based on the new input, ensuring the model's adaptability to sequential data.
	
	\item \textbf{Storing Updated Vectors:} The updated key-value vectors are then stored back into the cache, ready to be used in subsequent steps. This storage is essential for maintaining the continuity of the model's state across different time steps.
\end{enumerate}

This update mechanism is a cornerstone of the BCT, allowing it to harness the benefits of a fixed cache size while dynamically adjusting to new information, thus achieving a balance between computational efficiency and expressive power.

\subsection{Pseudocode}
\label{subsec:pseudocode}

The following pseudocode provides a detailed description of the key-value vector update process within our Bounded-Cache Transformer (BCT) layer:

\begin{algorithm}[H]
	\caption{Key-Value Vector Update Process}
	\label{alg:kv_update_process}
	\begin{algorithmic}[1]
		\Procedure{UpdateKV}{$\mathbf{x}_1, \ldots, \mathbf{x}_N$, $\mathbf{MV}_0$, $\mathbf{MK}$, $\mathbf{W}_{wq}$, $\mathbf{W}_{wv}$}
		\For{$t = 1$ to $N$}
		\State $\mathbf{wq}_t \gets \mathbf{W}_{wq} \cdot \mathbf{x}_t$ \Comment{Compute write query vector}
		\State $\mathbf{wv}_t \gets \mathbf{W}_{wv} \cdot \mathbf{x}_t$ \Comment{Compute write key-value vector}
		\State $\mathbf{score}_t \gets \mathbf{wq}_t \cdot \mathbf{MK}^T$ \Comment{Calculate similarity scores}
		\State $\mathbf{ww}_t \gets \text{softmax}(\mathbf{score}_t)$ \Comment{Normalize scores to weights}
		\If{$t = 1$}
		\State $\mathbf{MV}_t \gets \mathbf{MV}_0$ \Comment{Initialize with initial state}
		\Else
		\State $\mathbf{MV}_t \gets \mathbf{MV}_{t-1} + \mathbf{ww}_t^T \cdot \mathbf{wv}_t$ \Comment{Update key-value vector sequence}
		\EndIf
		\EndFor
		\State \Return $\mathbf{MV}_N$ \Comment{Return the final key-value vector sequence}
		\EndProcedure
	\end{algorithmic}
\end{algorithm}

\subsection{Illustration}
\begin{figure}[ht]
	\centering
	\includegraphics[width=0.8\textwidth]{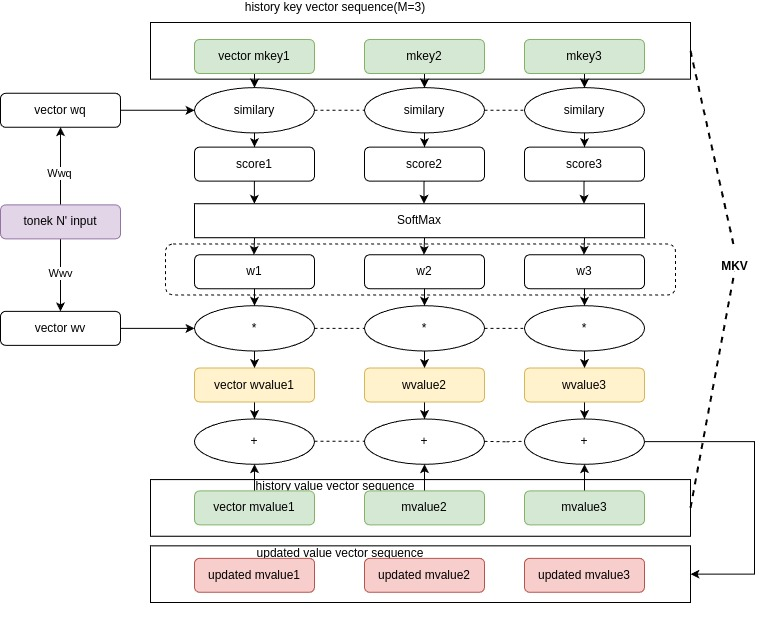}
	\caption{Illustration of the Key-Value Vector Update Process in BCT.}
	\label{fig:kv_update}
\end{figure}

To further clarify the key-value vector update mechanism within the Bounded-Cache Transformer (BCT), Figure \ref{fig:kv_update} offers a visual guide. This illustration encapsulates the core steps of the update process, which are as follows:

1. \textbf{Input Transformation:} The input vector $\mathbf{x}_t$ is transformed into a write query vector $\mathbf{wq}_t$ and a write key-value vector $\mathbf{wv}_t$ using the matrices $\mathbf{W}_{wq}$ and $\mathbf{W}_{wv}$.

2. \textbf{Attention Weights:} The write query vector $\mathbf{wq}_t$ interacts with the key vectors $\mathbf{MK}$ to compute attention weights $\mathbf{ww}_t$, which are then normalized using the softmax function.

3. \textbf{Cache Update:} The attention weights $\mathbf{ww}_t$ and the write key-value vector $\mathbf{wv}_t$ are used to update the key-value vector sequence in the cache. This update is mathematically represented by the formula:
\[
\mathbf{MV}_t = \mathbf{MV}_{t-1} + \mathbf{ww}_t^T \cdot \mathbf{wv}_t
\]
where $\mathbf{MV}_{t-1}$ is the previously stored key-value vector sequence.

4. \textbf{Cache Storage:} The updated key-value vectors are stored back into the cache, ready for the next input vector.

Figure \ref{fig:kv_update} provides a visual representation of these steps, highlighting the dynamic interaction between the input vectors, the key-value cache, and the attention mechanism within the BCT layer.

\subsection{Key Implementation Aspects}
\label{subsec:key_implementation}

The Bounded-Cache Transformer (BCT) introduces several key implementation aspects that enhance its efficiency and effectiveness in sequence modeling:

\begin{itemize}
	\item \textbf{Fixed-Length Cache:} BCT utilizes a fixed-length key-value (KV) cache, which contrasts with the expanding cache in traditional Transformers. This design is essential for managing memory efficiency and ensuring stable performance across varying context lengths.
	
	\item \textbf{Adaptive Update Rule:} BCT features an adaptive update rule that refines the hidden state through a self-supervised learning approach during inference. This enables the model to dynamically adjust to new sequences, improving its capacity to capture long-range dependencies within the bounded cache.
	
	\item \textbf{Optimized for Parallelism:} The BCT is designed to take full advantage of parallel processing capabilities. Since the update of KV content at each time step is independent, it allows for token-level parallel training. This is achieved through the computation of all KV elements via concurrent traversal and accumulation operations, significantly enhancing training efficiency.
	
	\item \textbf{Learning Rate Adaptation:} BCT incorporates an adaptive learning rate mechanism, which adjusts the step size for each token. This is vital for stabilizing the learning process and can lead to faster convergence.
	
	\item \textbf{Hardware Efficiency:} The BCT's design is mindful of modern hardware constraints, aligning its operations with the strengths of GPUs and TPUs. This includes the use of efficient matrix operations and strategic memory management to ensure high-performance computing.
\end{itemize}

These aspects highlight BCT's innovative approach to sequence modeling, focusing on efficiency, adaptability, and hardware utilization, which are critical for advancing the state of the art in language model design.

\section{Experimental Results}

\subsection{Datasets}
The experiments used the following datasets:

\textbf{Common Crawl}: A dataset containing a large amount of web content \cite{borg2019common}.
\textbf{RefinedWeb}: A web-crawled dataset that has been cleaned and filtered \cite{chung2022refinedweb}.
\textbf{The Pile}: A comprehensive dataset containing various types of text \cite{gao2020pile}.
\subsection{Experimental Setup}

\textbf{Model}: A large language model based on the Transformer architecture with 1B parameters.
\textbf{Cache Capacity}: Predefined as 2048.
\textbf{Evaluation Metrics}:
\textbf{Memory Usage}: Measure the memory consumed by the model during inference.
\textbf{Inference Speed}: Measure the time required to process each input vector.
\textbf{Inference Quality}: Evaluate the model's inference quality using BLEU scores.
\subsection{Results Analysis}

\textbf{Memory Usage}: After the inference length exceeds the fixed cache capacity (2048), the memory usage of the proposed method remains constant, significantly lower than that of traditional KV cache methods. Specifically, the memory usage of traditional methods increases linearly with context length, while the memory usage of the proposed method remains constant after reaching 2048.
\textbf{Inference Speed}: The inference speed also remains relatively constant after the inference length exceeds 2048. This indicates that the proposed method does not significantly degrade inference efficiency when handling long contexts.
\textbf{Inference Quality}: The BLEU scores are comparable to those of traditional methods, with differences within 3
\section{Discussion}
The proposed method for building large language models with predefined KV cache capacity dynamically updates key-value vector sequences, achieving efficient inference within limited cache capacity. This method maintains the model's inference quality while significantly reducing memory usage and improving system throughput. Future work will further explore the application of this method in larger parameter scales and larger training datasets, particularly in validating performance with longer context lengths. We plan to conduct experiments on models with larger parameter scales (e.g., 10B or more) and larger training datasets (e.g., more comprehensive web crawl data) to further verify the effectiveness and applicability of the method.

\section{Conclusion}
This paper proposes a method for building large language models with predefined KV cache capacity, addressing the issue of excessive memory consumption in traditional KV caches when handling infinite contexts. Experimental results show that the proposed method significantly reduces memory usage while maintaining the model's inference quality and improving system throughput. This method provides a new solution for efficient inference in large language models, offering important theoretical and practical significance.

\bibliographystyle{plain}
\bibliography{KV}

\end{document}